\documentclass[10pt,a4paper,twocolumn]{scrartcl}

\usepackage[utf8]{inputenc}
\usepackage{geometry}
\usepackage{ifthen}
\usepackage{amsmath}
\usepackage{pgfplotstable}
\usepackage{xfrac}
\usepackage{booktabs}       %
\usepackage{float}
\usepackage[]{algorithmic}
\usepackage[font={small},labelfont=bf]{caption}
\usepackage{authblk}
\usepackage{import}
\usepackage{pgfplotstable}
\usepackage{tikz}
\usepackage{hyperref}
\usepackage{soul}
\usepackage[notextcomp]{stix}
\usepackage{flushend}
\usepackage[]{todonotes}
\usepackage{multirow}
\usepackage{array}
\usepackage[roman]{parnotes}
 
\newif\ifarxiv
\arxivfalse

\newfloat{algorithm}{t}{lop}
\floatname{algorithm}{Algorithm}

\hypersetup{colorlinks,%
	citecolor=blue,%
	filecolor=black,%
	linkcolor=black,%
	urlcolor=black
}

\usetikzlibrary{shapes.misc,shapes}
\usetikzlibrary{decorations.pathreplacing}
\usetikzlibrary{positioning,fit,calc}
\tikzset{font=\sffamily}
\usetikzlibrary{shapes.misc,shapes}
\usetikzlibrary{decorations.pathreplacing}
\usetikzlibrary{positioning,fit,calc,spy}
\usetikzlibrary{arrows.meta,decorations.pathreplacing,decorations.text,fadings,shapes,arrows,patterns}
\usetikzlibrary{backgrounds}

\usetikzlibrary{arrows.meta,decorations.pathreplacing,fadings,shapes,arrows}

\definecolor{blue}{RGB}{0,91,130}
\definecolor{lightblue}{RGB}{110,159,189}
\definecolor{red}{RGB}{185,70,60}
\definecolor{lightred}{RGB}{198,141,132}
\definecolor{green}{RGB}{125,150,110}
\definecolor{lightgreen}{RGB}{164,181,153}
\definecolor{orange}{HTML}{D7AA50}
\definecolor{purple}{HTML}{7A68A6}

\definecolor{vblue}{HTML}{3b468d}
\definecolor{vyellow}{HTML}{fde900}
\definecolor{vgreen}{HTML}{35b779}

\definecolor{mblue}{HTML}{1f77b4}
\definecolor{mred}{HTML}{d62728}
\definecolor{morange}{HTML}{ff7f0e}

\definecolor{pred}{HTML}{e64c24}

\makeatletter
\def\relativepath{\import@path}
\makeatother

\setcapindent{0pt}

\deffootnote[0.3em]{0.0em}{0em}{\textsuperscript{\thefootnotemark}}

\makeatletter
\patchcmd{\@maketitle}{\LARGE \@title}{\fontsize{16}{19.2}\selectfont\@title}{}{}
\makeatother

\newcommand\addtabletext[1]{#1}

\setkomafont{title}{\normalfont\Huge}
\setkomafont{section}{\normalfont\Large}
\setkomafont{subsection}{\normalfont}
\setkomafont{paragraph}{\normalfont\bfseries}

\usepackage[numbers,sort&compress,merge,round]{natbib}
\makeatletter  \renewcommand\@biblabel[1]{#1.}  \makeatother

\usepackage{libertinus}
\usepackage{libertinust1math} %
\usepackage[scr=boondoxo,bb=boondox]{mathalfa} %
\renewcommand\sffamily{}

\usepackage[T1]{fontenc}

\usepackage{physics}
\usepackage[binary-units,detect-all,separate-uncertainty=true]{siunitx}
\DeclareSIUnit{\nothing}{\relax}
\DeclareSIUnit{\sample}{Sample}

\usepackage{glossaries}
\newacronym{ppu}{PPU}{plasticity processing unit}
\newacronym{padi}{PADI}{parallel driver interface}
\newacronym{aer}{AER}{address event representation}

\newacronym{sram}{SRAM}{static random-access memory}
\newacronym{adc}{ADC}{analog-to-digital converter}
\newacronym{cadc}{CADC}{column-parallel analog-to-digital converter}
\newacronym{dac}{DAC}{digital-to-analog converter}
\newacronym{dut}{DUT}{design under testing}
\newacronym{asic}{ASIC}{application-specific integrated circuit}
\newacronym{fpga}{FPGA}{field-programmable gate array}
\newacronym{vlsi}{VLSI}{very-large-scale integration}
\newacronym{simd}{SIMD}{single instruction, multiple data}
\newacronym{ocp}{OCP}{Open Core Protocol}
\newacronym{soc}{SoC}{system on a chip}
\newacronym{sta}{STA}{static timing analysis}
\newacronym{gals}{GALS}{globally asynchronous locally synchronous}
\newacronym{dpi}{DPI}{SystemVerilog direct programming interface}
\newacronym{pll}{PLL}{phase-locked loop}
\newacronym{fifo}{FIFO}{First-In-First-Out Buffer}

\newacronym{psc}{PSC}{post-synaptic current}
\newacronym{psp}{PSP}{post-synaptic potential}
\newacronym{stp}{STP}{short-term plasticity}
\newacronym{stdp}{STDP}{spike-timing-dependent plasticity}
\newacronym{rstdp}{R-STDP}{reward-modulated spike-timing-dependent plasticity}
\newacronym{lif}{LIF}{leaky integrate-and-fire}
\newacronym{adex}{AdEx}{adaptive exponential leaky integrate-and-fire}
\newacronym{snn}{SNN}{spiking neural network}
\newacronym{dnn}{DNN}{deep neural network}
\newacronym{mlp}{MLP}{multi-layer perceptron}
\newacronym{ann}{ANN}{artifical neural network}

\newacronym{mc}{MC}{Monte Carlo}

\newacronym{bptt}{BPTT}{backpropagation through time}
\newacronym{gpu}{GPU}{graphics processing unit}
\newacronym{relu}{ReLU}{rectified linear unit}

\newacronym{itl}{ITL}{in-the-loop}

\newacronym{shd}{SHD}{spiking Heidelberg digits}

\title{Surrogate gradients for analog neuromorphic computing}

\author[a,1,2]{B.~Cramer}
\author[a,1,2]{S.~Billaudelle}
\author[a]{\\S.~Kanya}
\author[a]{A.~Leibfried}
\author[a]{A.~Grübl}
\author[a]{V.~Karasenko}
\author[a]{C.~Pehle}
\author[a]{K.~Schreiber}
\author[a]{Y.~Stradmann}
\author[a]{J.~Weis}
\author[a]{\\J.~Schemmel}
\author[b,2]{F.~Zenke}

\affil[a]{Kirchhoff-Institute for Physics, Heidelberg University, Germany}
\affil[b]{Friedrich Miescher Institute for Biomedical Research, Basel, Switzerland}
\affil[1]{Authors with equal contribution}
\affil[2]{Corresponding authors (benjamin.cramer,sebastian.billaudelle@kip.uni-heidelberg.de,friedemann.zenke@fmi.ch)}

\date{}

\begin{document}

\maketitle

\begin{abstract}
\noindent\bfseries
To rapidly process temporal information at a low metabolic cost, biological neurons integrate inputs as an analog sum but communicate with spikes, binary events in time. 
Analog neuromorphic hardware uses the same principles to emulate spiking neural networks with exceptional energy-efficiency.
However, instantiating high-performing spiking networks on such hardware
remains a significant challenge due to device mismatch and the lack of efficient training algorithms.
Here, we introduce a general in-the-loop learning framework based on surrogate gradients that resolves these issues.
Using the {BrainScaleS-2} neuromorphic system, we show that learning
self-corrects for device mismatch resulting in competitive spiking network performance on both vision and speech benchmarks.
Our networks display sparse spiking activity with, on average, far less than one spike per hidden neuron and input, perform inference at rates of up to \SI{85}{\kilo\nothing} frames/second, and consume less than \SI{200}{\milli\watt}.
In summary, our work sets several new benchmarks for low-energy spiking network processing on analog neuromorphic hardware and paves the way for future on-chip learning algorithms.

\end{abstract}
\vspace{0.5em}

{
\bfseries
\itshape
\noindent
}

\glsresetall

\section*{Introduction}
In recent years, deep \glspl{ann} have surpassed human-level performance on many difficult tasks \citep{mnih2013playing, silver2017mastering, brown2020language}.
The human brain, however, remains unchallenged in terms of its energy-efficiency and fault tolerance.  
A fundamental property underlying these capabilities is spatiotemporal sparseness \citep{sterling2015principles}, which is directly linked to the way how biological \glspl{snn} process and exchange information. 
Spiking neurons receive and integrate inputs on their analog membrane potentials and, upon reaching the firing threshold, emit action potentials, or spikes.
These binary events propagate asynchronously through the \gls{snn} and are ultimately received by other neurons. 

Neuromorphic engineering attempts to mirror the power efficiency and robustness of the brain by replicating its key architectural properties \citep{thakur2018large, schuman2017survey}.
Here, one distinguishes between fully digital, analog, and mixed-signal systems.
Digital systems \emph{simulate} the analog dynamics of spiking neurons, e.g., their membrane potentials \citep{merolla2014million, benjamin2014neurogrid, furber2016large, boahen2017neuromorphs, davies2018loihi, roy2019towards}.
In contrast, analog and mixed-signal solutions \emph{emulate} neuronal or synaptic dynamics and states by representing them as physical voltages, currents, or conductance changes evolving in continuous time \citep{mahowald1991silicon, schemmeliscas2010, indiveri2011neuromorphic, chicca2014neuromorphic}.
Thus, by explicitly taking advantage of physical properties and dynamics of the underlying hardware substrate, neuromorphic computing holds the key to building power-efficient and scalable \glspl{snn} in-silico \cite{roy2019towards, markovic2020physics, ambrogio2018equivalent}.

However, to serve meaningful computational purpose, these analog devices require training.
The most successful training schemes for \glspl{ann} are gradient-based.
Yet, extending similar training techniques to \glspl{snn} and neuromorphic hardware poses several challenges.
First, one has to overcome the binary nature of spikes, which impedes vanilla gradient-descent \citep{bengio2013estimating, courbariaux2016binarized, neftci2019surrogate}.
Second, training has to ensure sparse spiking activity to exploit the superior power efficiency of \gls{snn} processing \citep{davidson2021comparison}.
Finally, training has to achieve all of the above while coping with analog hardware imperfections inevitably tied to their manufacturing process.

In this article, we tackle the above challenges by extending previous work on surrogate gradients.
Specifically, we developed a general \gls{itl} training framework and applied it to the mixed-signal BrainScaleS-2 single-chip system.
We demonstrate that \glspl{snn} trained using our approach solve several challenging benchmark problems by taking advantage of sparse, precisely timed spikes instead of firing rates.
The resulting \glspl{snn} reach competitive accuracy levels comparable to corresponding software simulations and perform inference with ultra-low latency by taking full advantage of BrainScaleS' hardware acceleration.
Crucially, we show that \gls{itl} surrogate gradients achieve this through self-calibration, whereby training automatically corrects for device mismatch without the need for costly offline calibration.

\section*{The BrainScaleS-2 analog neuromorphic substrate}

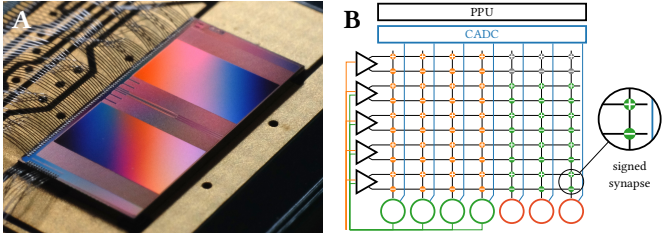
\begin{figure}[t!]
	\centering
	\begin{tikzpicture}[]
		\node[anchor=north west,inner sep=0pt,draw=none] (a) at (0.0,0) {\includegraphics[width=4.2cm]{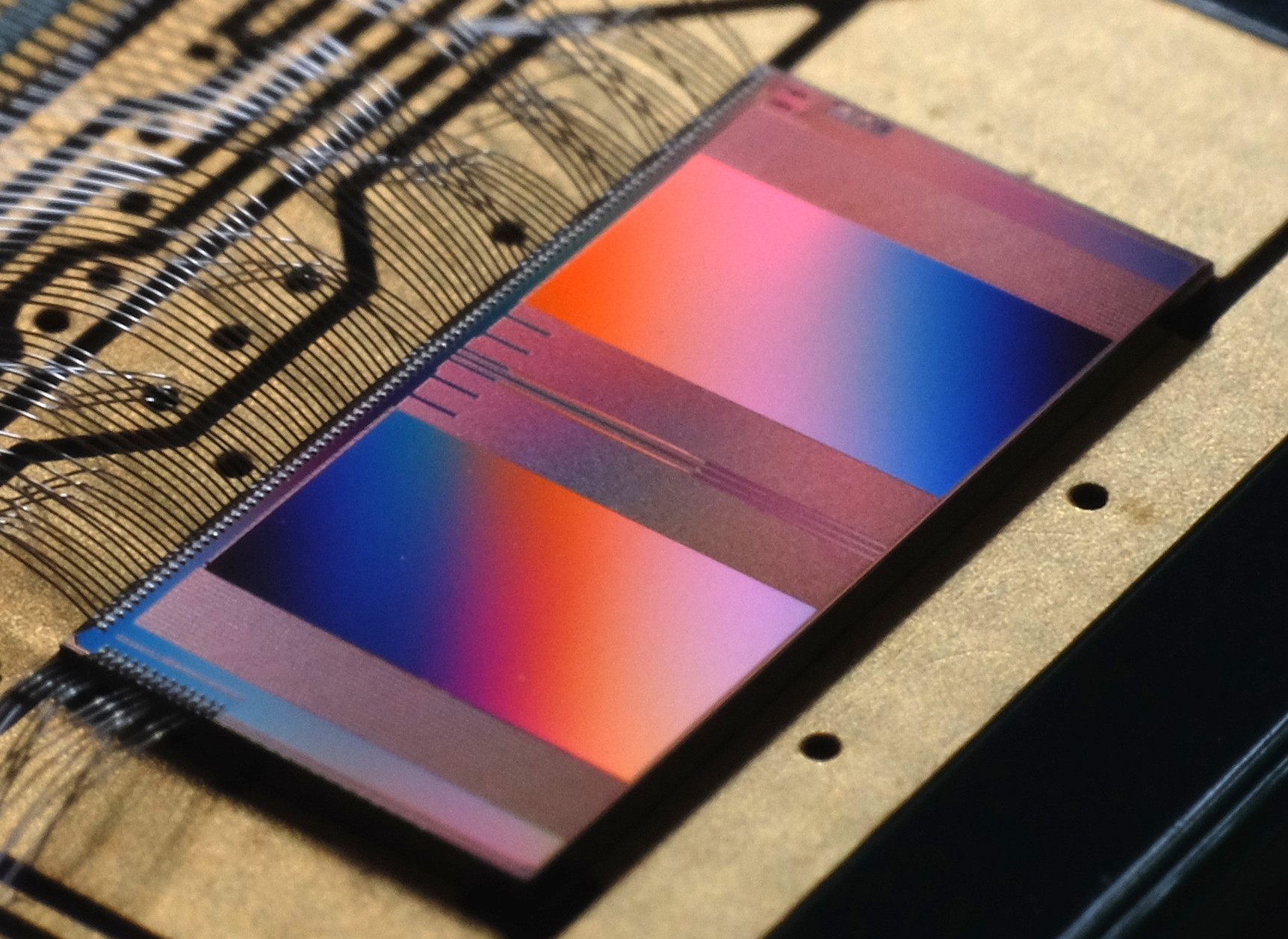}};
		\node[anchor=north west,inner sep=0pt,draw=none] (b) at (4.5,0) {\scalebox{0.85}{\usetikzlibrary{spy}%
\definecolor{blue}{HTML}{1f77b4}%
\definecolor{red}{HTML}{d62728}%
\definecolor{green}{HTML}{2ca02c}%
\definecolor{orange}{HTML}{ff710e}%
\definecolor{yellow}{HTML}{fee23e}%
\definecolor{cadc}{HTML}{1f77b4}%
\definecolor{input}{HTML}{ff7f0e}%
\definecolor{hidden}{HTML}{2ca02c}%
\definecolor{output}{HTML}{555555}%
\tikzset{block/.style={font={\rmfamily\footnotesize},align=center}}%
\tikzset{box/.style={draw=black!90}}%
\tikzset{block label/.style={fill=white,font={\rmfamily\footnotesize},inner sep=0.05cm}}%
\tikzset{%
	neuron/.style = {%
		draw=black,%
		circle,%
		inner sep=0pt,%
		minimum width=0.4cm%
	},%
	driver/.style = {%
		minimum height=0.45cm,%
		draw=black,%
		regular polygon,%
		regular polygon sides=3,%
		shape border rotate=-90,%
		inner sep=0pt%
	},%
}%
\begin{tikzpicture}[
            scale=0.8,
            >=stealth,
            transform shape,
	    line width=1.0\pgflinewidth,
	    anchor=center,
	    spy using outlines=circle,
        ]
        \pgfdeclarelayer{background layer}
        \pgfsetlayers{background layer,main}
        \draw[use as bounding box,inner sep=0pt,draw=none] (0.0,0.0) rectangle ++(7.0,4.5);

	\begin{scope}[scale=1.15]
		\foreach \x in {0,1,...,6} {
			\pgfmathparse{100*(\x>3)}
			\colorlet{currentcolor}{pred!\pgfmathresult!hidden}
			
			\node[neuron,currentcolor,thick] (nrn \x) at (0.8 + \x*0.5,0.35) {};
			\draw (nrn \x.north) ++ (0.0,0.01) -- ++(0.0,2.5);
		}

		\foreach \y in {0,1,...,4} {
			\node[driver,thick] (drv \y) at ($(nrn 0) + (-0.5,0.5 + \y*0.5)$) {};
			\draw (drv \y.+45) -- ($(drv \y.center) + (0.15, 0.125)$) -- ++(3.5,0.0);
			\draw (drv \y.-45) -- ($(drv \y.center) + (0.15,-0.125)$) -- ++(3.5,0.0);

			\foreach \x in {0,1,...,6} {
				\pgfmathparse{100*(\x>3)}
				\colorlet{currentcolor}{hidden!\pgfmathresult!input}
				\pgfmathparse{100*(\x>3)*(\y>3)}
				\colorlet{currentcolor}{gray!\pgfmathresult!currentcolor}

				\fill[currentcolor] (drv \y -| nrn \x) ++ (0.0, 0.125) circle (0.05cm);
				\draw[white] (drv \y -| nrn \x) ++ (0.0, 0.125) ++ (-0.045,0.0) -- ++(0.09,0.0);
				\draw[white] (drv \y -| nrn \x) ++ (0.0, 0.125) ++ (0.0,-0.045) -- ++(0.0,0.09);
				\fill[currentcolor] (drv \y -| nrn \x) ++ (0.0,-0.125) circle (0.05cm);
				\draw[white] (drv \y -| nrn \x) ++ (0.0,-0.125) ++ (-0.045,0.0) -- ++(0.09,0.0);
			}
		}

		\foreach \x in {0,1,2,3} {
			\draw[hidden] (nrn \x.south) -- ++(0.0,-0.1) coordinate (tmp) -- (tmp -| drv 0.west) -- ++(-0.1,0.0) coordinate (sammelpunkt);
		}
		
		\foreach \y in {0,1,...,3} {
			\draw[hidden] (sammelpunkt) -- ($(sammelpunkt |- drv \y) - (0.0,0.03)$) coordinate (tmp) -- (tmp -| drv \y.west);
		}
		
		\foreach \y in {0,1,...,4} {
			\draw[input] ($(sammelpunkt) - (0.06,0.0)$) coordinate (tmp) -- ($(tmp |- drv \y) + (0.0,0.03)$) coordinate (tmp) -- (tmp -| drv \y.west);
		}

		\node[rectangle,thick,draw,cadc,inner sep=2pt,minimum width=3.5cm] (cadc) at ($(nrn 2) + (0.5,3.0)$) {\fontsize{6}{6}\selectfont CADC};
		\foreach \x in {0,1,...,6} {
			\draw[blue] (nrn \x.55) -- ++(55:0.12) coordinate (tmp) -- (tmp |- cadc.south);
		}
		
		\node[rectangle,thick,draw,inner sep=2pt,minimum width=3.5cm,above=0.07cm of cadc] (ppu) {\fontsize{6}{6}\selectfont PPU};

	\end{scope}

	\coordinate (sherlock) at (5.5,2.2);
	\spy[draw,height=1.2cm,width=1.2cm,magnification=2.5,connect spies] on (nrn 6 |- drv 0) in node at (sherlock);
	\node[align=center] at ($(sherlock) - (0.0,1.1)$) {\fontsize{7}{7}\selectfont signed\\[-0.3em] \fontsize{7}{7}\selectfont synapse};
\end{tikzpicture}}};

		\node[inner sep=0pt,white] at ($(a.north west) + ( 0.25,-0.25)$) {\bfseries A};
		\node[inner sep=0pt,black] at ($(b.north west) + (0.10,-0.25)$) {\bfseries B};
	\end{tikzpicture}
	\caption{
		\textbf{The mixed-signal BrainScaleS-2 chip.}
	    \textbf{(A)} Close-up chip photograph.
	    \textbf{(B)} Implementation of a multi-layer network on the analog neuromorpic core.
	    Input spike trains are injected via synapse drivers (triangles) and relayed to the hidden layer neurons (green circles) via the synapse array.
	    Spikes in the hidden layer are routed on-chip to the output units (red circles).
	    Each connection is represented by a pair of excitatory and inhibitory
		hardware synapses, which holds a signed weight value.
	    The analog membrane potentials are read out via the \gls{cadc} and further processed by the \gls{ppu}.
	    }
	\label{fig:arch}
	\glsreset{ppu}
	\glsreset{cadc}
\end{figure}

In this article we relied on the analog BrainScaleS-2 single chip system.
It features 512 analog neuron circuits whose dynamics obey the \gls{lif} equation
\begin{equation}
	C \dv{V}{t} = -g_\text{leak} \left( V - V_\text{leak} \right) + I \,,
	\label{eqn:lif}
\end{equation}
which can optionally be augmented by adaptation currents and an exponential spiking nonlinearity.
The membrane potential $V$ is explicitly represented on the chip as an analog voltage measured across a capacitor and evolves continuously in time.
The leak conductance $g_\text{leak}$ pulls the membrane towards the leak potential $V_\text{leak}$, resulting in an exponential decay with time constant $\tau_\text{m} \equiv C/g_\text{leak}$.
Due to the substrate's small intrinsic capacitances and comparatively large currents, the dynamics of the spiking neurons implemented on BrainScaleS-2 evolve 10\textsuperscript{3} times faster than biological neurons.

Whenever the potential crosses the firing threshold $\vartheta$, an outgoing spike is generated and the membrane is reset.
An on-chip event router propagates both internally generated and external spikes to connected neurons, allowing to form feed-forward as well as recurrent topologies.
To that end, each neuron integrates stimuli from a column of 256 synapses, where the weights are represented with a resolution of \SI{6}{\bit}.
The resulting postsynaptic currents $I$, which are integrated on the membrane capacitor, follow an exponential time course similar to the membrane dynamics themselves.
The sign of a synapse is determined as a presynaptic property.
However, we allowed for a continuous transition between positive and negative weights during training by merging synapse circuits of opposing signs (Fig.~\ref{fig:arch}B).

BrainScaleS-2 allows individually adjusting all neuronal parameters, including reference potentials and time constants, on a per-neuron basis to flexibly emulate different target dynamics.
This fine-grained control also facilitates calibration to mitigate deviations induced by variations in the production process.
In this article, we however make use of this parameterization to actively \emph{decalibrate} the system, thereby allowing us to systematically explore self-calibration properties of our learning algorithm.

\section*{In-the-loop surrogate gradients on analog hardware}

\begin{figure}[t!]
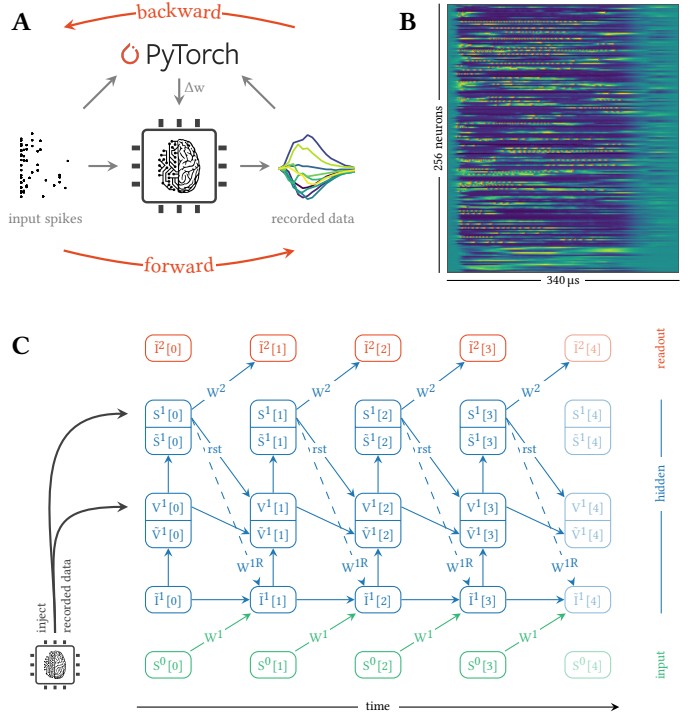

	\centering
	\begin{tikzpicture}[]
		\node[anchor=north west,inner sep=0pt] (a) at (0,4.3) {\subimport{figures/flow/}{flow.tex}};
		\node[anchor=north east,inner sep=0pt] (b) at (\columnwidth,4.3) {\begin{tikzpicture}
	\node (traces) at (0, 0) {\input{traces.pgf}};
	
	\begin{scope}[on background layer]
	\draw ($(traces.south west) + (0, -0.1cm)$) -- ($(traces.south east) + (0, -0.1cm)$) node[pos=0.5, fill=white, anchor=center, inner sep=2pt] {\tiny\vphantom{gl}\SI{340}{\micro\second}};
	\draw ($(traces.north west) + (-0.1cm, 0)$) -- ($(traces.south west) + (-0.1cm, 0)$) node[pos=0.5, fill=white, anchor=center, rotate=90, inner sep=2pt] {\tiny \vphantom{gl}256 neurons};
	\end{scope}
\end{tikzpicture}};
		\node[anchor=north,inner sep=0pt] (c) at (0.5\columnwidth,0) {\subimport{figures/graph/}{graph.tex}};

		\node[inner sep=0pt] at ($(a.north west) + (0.25,-0.25)$) {\bfseries A};
		\node[inner sep=0pt] at ($(b.north west) + (-0.25,-0.25)$) {\bfseries B};
		\node[inner sep=0pt] at ($(c.north west -| a.west) + (0.25,-0.25)$) {\bfseries C};

	\end{tikzpicture}
	\caption{
	    \textbf{Surrogate gradient learning on BrainScaleS-2.}
	    \textbf{(A)} Illustration of our \gls{itl} training scheme.
	    The forward pass is emulated on the BrainScaleS-2 chip.
	    Observables from the neuromorphic substrate as well as the input spike trains are processed on a conventional computer to perform the backward pass.
	    The calculated weight updates are then written to the neuromorphic system.
	    \textbf{(B)} Parallel recording of analog traces and spikes from 256 neurons via the \acrshort{cadc}.
	    \textbf{(C)} The differentiable computation graph results from the integration of \acrshort{lif} dynamics.
	    	The time dimension is unrolled from left to right and information flows from bottom to top within an integration step.
		Synaptic currents are derived from the previous layer's spikes and potential recurrent connections, multiplied by the respective weights (W\textsuperscript{(R)}).
		Stimuli are integrated on the neurons' membranes (V) which trigger spikes (S) upon crossing their thresholds.
		These observables are continuously synchronized with data recorded from the hardware.
		Spikes as well as reset signals (rst) are propagated to the next time step, which also factors in the decay of currents and potentials.
	    }
	\label{fig:itl}
	\glsreset{ppu}
	\glsreset{cadc}
\end{figure}

To train \glspl{snn} on BrainScaleS-2, we developed a general learning framework to optimize recurrent and multi-layer networks.
Our approach is based on the notion of surrogate gradients, which overcome vanishing gradients and critical points associated with non-differentiable spiking dynamics \citep{neftci2019surrogate}.
Surrogate gradient learning flexibly supports arbitrary differentiable loss functions and can seamlessly exploit both rate- and spike timing-based coding schemes.

Broadly, our \gls{itl} approach works as follows (Fig.~\ref{fig:itl}):
First, we emulate the forward pass on the analog neuromorphic substrate and record both spikes and internal membrane traces (Fig.~\ref{fig:itl}A,B).
By injecting the latter into an otherwise approximate software model we effectively render the neuromorphic \gls{snn} differentiable.
This permits the evaluation of surrogate gradients and the calculation of weight updates using \gls{bptt} on GPU-enabled auto-differentiation libraries \citep{NEURIPS2019_9015} in combination with state-of-the-art optimizers.
At the same time, our learning algorithm self-corrects for parameter mismatch of the analog components (Fig.~\ref{fig:itl}C).
Finally, we close the loop by transferring the updated weights back to the analog system.

In the following we elaborate on the two central steps, namely the recording of data from the neuromorphic system and their integration into the computation graph.

\paragraph{Recording spikes and analog membrane traces.}
Surrogate gradient learning crucially relies on the neurons' membrane potentials.
On an analog system like BrainScaleS-2, these are represented as physical voltages and are hence not readily available for numerical computation.
The required digitization is often challenging due to the inherent parallelism of these substrates.
This bottleneck is further emphasized by accelerated systems.

BrainScaleS-2 solves this problem by incorporating \glspl{cadc} to simultaneously digitize the membrane potentials of all neurons (Fig.~\ref{fig:arch}B).
We trigger the \acrshort{adc} conversions via the embedded \glspl{ppu} \citep{friedmann2016hybridlearning} to ensure higher and more stable sampling rates compared to a host-based scheduling.
This furthermore enables the implementation of a fast inference mode, where only classification results are transmitted to the host.
When training the network, however, each recorded sample is instantly transferred to an intermediate external memory region, from where it is asynchronously read by the host machine at the end of an input pattern or batch.
In total we reach a sample rate of approximately \SI{0.6}{\mega\sample\per\second}, corresponding to a sampling interval of \SI{1.7}{\micro\second}.
For 256 neurons, this yields a total data rate of \SI{1.2}{\giga\bit\per\second}.
In addition to the sampled membrane traces, we continuously record and time stamp the spike events emitted by the substrate.

\paragraph{A computation graph for analog circuits.}
To compute weight updates based on surrogate gradients, we incorporate these aggregated data into a computation graph that approximates the underlying neuronal dynamics on the neuromorphic substrate. 
To that end, we iteratively simulate the neuronal dynamics to obtain the graph in which we inject the \emph{actual} recorded membrane traces.
Thus we use measured quantities, where available, and only rely on the model \emph{estimates} for internal variables that are not measured, e.g., the synaptic currents.

We formulate the graph on a regular time grid of time step $\Delta t$ derived from the sampling period of the membrane traces.
Although the spike trains from the neuromorphic substrate are known with much higher temporal resolution, they are also aligned to these bins.
Depending on the coding scheme and network topology, an increased resolution can be beneficial and allow to better capture causal relations between spikes.
In this case, the computation graph can be evaluated on a finer time scale and for that purpose operate on interpolated membrane traces.

To reconstruct the internal states we start by assuming ideal \gls{lif} dynamics (Eq.~\ref{eqn:lif}), which we integrate using the forward Euler method.
The membrane evolution is estimated recursively by taking into account its temporal decay and the calculated synaptic currents $\tilde I[t]$, which in turn are based on the presynaptic spikes $\tilde S_j[t]$ of neuron $j$:
\begin{align}
	\tilde V[t+1] &= \tilde V[t] \cdot \operatorname{e}^{-\Delta t/\tau_\text{m}} + \tilde I[t] \,, \label{eqn:membrane_forward} \\
	\tilde I[t+1] &= \tilde I[t] \cdot \operatorname{e}^{-\Delta t/\tau_\text{s}} + \textstyle\sum_j W_{j} \tilde S_j[t] \label{eqn:currents} \,. 
\end{align}
Eq.~\ref{eqn:currents} can be augmented by an additional term to encompass recurrent connections.
The modelled state variables, indicated by the tilde ($\,\tilde\,\,$), represent the estimates of the on-chip dynamics.
Since these can deviate from the actual emulation and hence distort the resulting gradients, we in their place insert the normalized recorded data.
For this purpose, we introduce an auxiliary identity function $f(x,\tilde x) \equiv x$ and define surrogate derivatives $\partial f/\partial x = 0$ and $\partial f/\partial \tilde x = 1$.
Eq.~\ref{eqn:membrane_forward} can now be modified to
 \begin{align}
	 \tilde V[t+1] &= f\left( V[t+1], \tilde V[t] \cdot \operatorname{e}^{-\Delta t/\tau_\text{m}} + \tilde I[t] \right) \,. \label{eqn:membrane_forward_mod}
\end{align}
A similar approach is taken for spikes by defining $\tilde S_j[t](S_j[t],\tilde V_j[t]) \equiv S_j[t]$ with associated derivatives
\begin{align}
	\frac{\partial \tilde S_j[t]}{\partial S_j[t]} &= 0 \,, &
	\frac{\partial \tilde S_j[t]}{\partial \tilde V_j[t]} &= \left( \beta \cdot |\tilde V_j[t] - \vartheta| \right)^{-2} \,, \label{eqn:spike-surrogate}
\end{align}
where $\beta$ describes the steepness of the surrogate gradient~\citep{zenke2018superspike}.

When performing the backward pass and to this end calculating $\partial \mathcal{L}/\partial \theta = \ldots \partial \tilde S / \partial \tilde V \cdot \partial \tilde V / \partial \theta$, the sampled values for the membrane potential are used whenever an expression containing $\tilde V$ is evaluated, e.g. in $\partial \tilde S / \partial \tilde V$.
The estimates, in contrast, are used to determine further derivatives $\partial\tilde V / \partial \theta$ which occur in the recursion relation of \gls{bptt}.

\paragraph{Flexible choice of a loss function.}
The suggested framework allows to operate on any differentiable loss that can be formulated on the data acquired from the neuromorpic system.
This encompasses loss functions based on the spiking activity of the neurons as well as on their membrane voltages (cf.~\nameref{sec:methods}).

The task-specific loss can be augmented by regularization functions.
These might, on one hand, target an improved generalization performance or, on the other hand, an adaptation to hardware-specific constraints such as finite weights and dynamic ranges of analog signals.
Such terms can furthermore be directly tailored to shape the activity of the emulated \glspl{snn} and result in sparse firing patterns.

\section*{Results}

\begin{figure*}[tbhp]
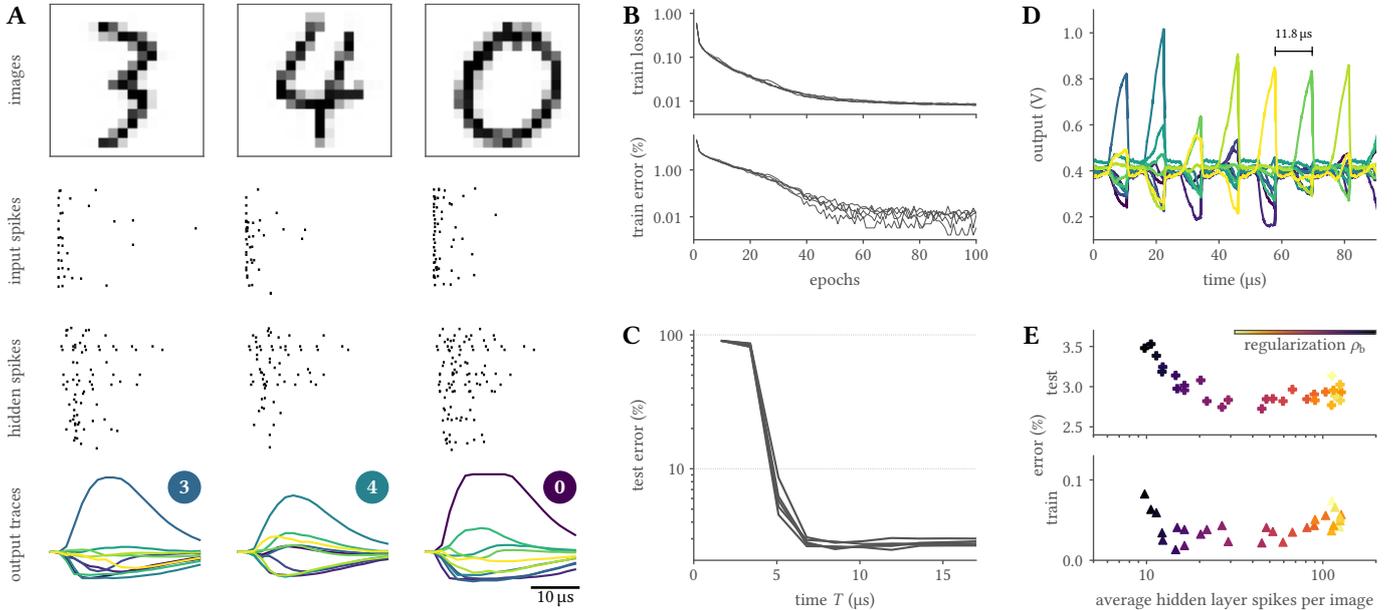
%
		\begin{tikzpicture}
			\node[anchor=north west,inner sep=0pt] (a) at (0.0,0) {\subimport{figures/spikes_and_traces/}{spikes_and_traces.pgf}};
			\node[anchor=north west,inner sep=0pt] (b) at (8.1,0) {\input{figures/accuracy/accuracy.pgf}};
			\node[anchor=north west,inner sep=0pt] (c) at (8.1,-4.25) {\input{figures/latency/latency.pgf}};
			\node[anchor=north east,inner sep=0pt] (d) at (\textwidth,0) {\subimport*{figures/superfast/}{traces.pgf}};
			\node[anchor=north east,inner sep=0pt] (e) at (\textwidth,-4.25) {\subimport*{figures/regularization/}{regularization.pgf}};
			
			\node[inner sep=0pt] at ($(a.north west) + (0.11,-0.15)$) {\bfseries A};
			\node[inner sep=0pt] at ($(b.north west) + (0.11,-0.15)$) {\bfseries B};
			\node[inner sep=0pt] at ($(c.north west) + (0.11,-0.15)$) {\bfseries C};
			\node[inner sep=0pt] at ($(d.north west) + (0.11,-0.15)$) {\bfseries D};
			\node[inner sep=0pt] at ($(e.north west) + (0.11,-0.15)$) {\bfseries E};
		\end{tikzpicture}
	\caption{
		\textbf{Classification of the MNIST dataset.}
		\textbf{(A)} Three snapshots of the \gls{snn} activity, consisting of the downscaled 16$\times$16 input images (top), spike raster of both the input spike trains and hidden layer activity (middle), and readout neuron traces (bottom).
			The latter show a clear separation, and hence a correct classification of the presented images.
		\textbf{(B)} Loss and accuracy over the course of 100 training epochs for five initial conditions.
		\textbf{(C)} The time to decision is consistently below \SI{10}{\micro\second}.
			Here, the classification latency was determined by iteratively re-evaluating the max-over-time for output traces (see panel A) restricted to a limited interval $[0,T]$.
		\textbf{(D)} This low latency allowed to inject an image every \SI{11.8}{\micro\second}, corresponding to more than \SI{85}{\kilo\nothing} classifications per second.
			This was achieved by artificially resetting the state of the neuromorphic network in between samples.
		\textbf{(E)} The neuromorphic system can be trained to perform classification with sparse activity.
			When sweeping the regularization strength, a state of high performance was evidenced over more than an order of magnitude of hidden layer spike counts.
	    }
	\label{fig:mnist_traces}
\end{figure*}

We trained BrainScaleS-2 on a series of spike-based vision and speech recognition tasks using our \gls{itl} learning framework.
Specifically, we chose classification tasks requiring evidence integration on widely different time scales which allowed us to probe the efficiency of our approach on both feed-forward or recurrent network topologies.

First, we trained a feed-forward network consisting of a single hidden layer with 246 neurons on the MNIST dataset~\citep{lecun1998gradient}.
To accommodate the data to a fan-in of 256~inputs, we reduced the original $28\times28$ images to 16$\times$16 pixels.
We then converted the pixels into a spike-latency code (cf.~\nameref{sec:methods}).
The network was optimized using the Adam optimizer~\citep{kingma2014adam} to minimize a max-over-time loss $ \mathcal{L} = \operatorname{NLL} ( \operatorname{softmax} ( \operatorname{max}_t \, V_i^\text{O}[t] ), y^\star ) $, with the negative log-likelyhood NLL, the membrane traces of the output layer $V_i^\text{O}[t]$, and the true labels $y^\star$.
To prevent excessive amplitudes and in turn clipping of $V_i^\text{O}$ on the analog substrate, we included a penalty $\rho_\text{a} \cdot \operatorname{mean}_i ( (\max_t V_i^\text{O}[t] )^2)$.
We furthermore added an activity shaping term to promote sparse activity patterns (cf.~Eq.~\ref{eqn:burst_reg}).
Notably, this contribution could only reduce the network's activity and did not act as an upwards pulling homeostatic force.
Being based on surrogate gradients, our approach nevertheless allowed training the network starting from a quiescent hidden layer.

During training, the neuromorphic substrate learned to correctly infer and represent the correct class memberships as the maximally responsive output units (Fig.~\ref{fig:mnist_traces}A,B).
Interestingly, the inhibition of the other units was not explicitly demanded by the loss function but emerged naturally through optimization.
After 100 epochs, the model almost perfectly fit the training samples and achieved an overall accuracy of \SI{97.2 \pm 0.1}{\percent} on held out test data (Table~\ref{tab:results}).
We were able to reduce overfitting by augmenting the data through random rotations of up to \SI{15}{\degree}.
Dropout similarly improved test performance and combining it with data augmentation resulted in an accuracy of \SI{97.6\pm0.1}{\percent} on BrainScaleS-2. %

\begin{table}[t!]
	\fontsize{8.5}{10}\selectfont
	\centering
	\caption{
		Comparison of results achieved with networks trained on BrainScaleS-2 and in software as well as an \acrshort{ann} baseline.
		}
	\label{tab:results}
	\newcommand{\hlr}{}

	\begin{tabular}{cllrr}
		\toprule
		        & implementation & remarks & \multicolumn{2}{c}{accuracy (\si{\percent})} \\
			&           &         & \multicolumn{1}{c}{train} & \multicolumn{1}{c}{test} \\
		\midrule
		\multirow{5}{*}{\rotatebox[origin=c]{90}{16$\times$16 MNIST}}
			& BSS-2       	&               	&  \num{100.0 \pm 0.0}	& \num{97.2 \pm 0.1} \\
			& BSS-2       	& dropout + rotation 	&  \num{97.3 \pm 0.1}	& \num{97.6 \pm 0.1} \\
			& \hlr software &			&  \hlr \num{100.0 \pm 0.0}  & \hlr \num{97.5 \pm 0.1} \\
			& \hlr software & dropout + rotation	&  \hlr \num{97.7 \pm 0.1}  & \hlr \num{98.0 \pm 0.0} \\
			& \hlr reference ANN	&			&  \hlr \num{100.0 \pm 0.0}  & \hlr \num{98.1 \pm 0.1} \\
			& \hlr reference ANN	& \hlr dropout + rotation 	&  \hlr \num{99.0 \pm 0.0}	& \hlr \num{98.7 \pm 0.1} \\
		\midrule
		\multirow{4}{*}{\rotatebox[origin=c]{90}{SHD}}
			& BSS-2 	&       	        &  \num{96.6 \pm 0.5}	& \num{76.2 \pm 1.3} \\
			& BSS-2 	& augmentation 	        &  \num{90.7 \pm 0.5}	& \num{80.6 \pm 1.0} \\
			& software 	&			&  \num{100.0 \pm 0.0}  & \num{71.2 \pm 0.3} \\
			& software 	& augmentation		&  \num{90.9 \pm 0.2}  & \num{79.9 \pm 0.7} \\
		\bottomrule
	\end{tabular}
\end{table}

As a comparison, we trained the same \gls{snn} purely in software and in that process ignored all hardware-specific constraints, including the finite weight resolution.
With an accuracy of \SI{97.5\pm0.1}{\percent} on the test data, the software implementation only slightly surpassed BrainScaleS-2.
As a baseline for the downscaled 16$\times$16 MNIST dataset, we furthermore trained an equivalently sized \gls{ann} with \glspl{relu} which resulted in an accuracy of \SI{98.1\pm0.1}{\percent}.
Dropout as well as augmentation again improved upon these numbers resulting in a best-effort performance of \SI{98.7\pm0.1}{\percent}.
Importantly, these accuracy figures -- within their uncertainties -- resembled results on the full-size MNIST images, suggesting comparability between these two datasets.

\paragraph{Low-latency neuromorphic computation.}
The output traces of trained networks suggested that for latency-encoded inputs as used above the decision is available long before the end of a stimulus (cf.~Fig.~\ref{fig:mnist_traces}A).
To determine the network's classification latency, we artificially restricted the readout layer's membrane traces (cf.~Fig.~\ref{fig:mnist_traces}A) to varying time intervals $[0, T]$ over which we based the network's decision as given by the maximally active unit.
We found that the readout reached its peak accuracy within \SI{8}{\micro\second} after the first input spike (Fig.~\ref{fig:mnist_traces}C).

Low classification latency, however, does not automatically translate into high inference rates but is also affected by the neuronal and synaptic time constants.
These time constants determine the rate by which state variable decay back to baseline within the neuron circuits and, hence, impose a minimum separation of independent stimuli.
Still, to translate low classification latency into high inference rates, we added an artificial reset of the neuromorphic units \SI{10}{\micro\second} after inserting the first input spike.
Specifically, we exploited a feature of BrainScaleS-2 that allowed us to concurrently reset the analog membrane circuits and clamped all synaptic currents to their respective baselines (Fig.~\ref{fig:mnist_traces}D).
This allowed us to infer images with a separation of \SI{11.8}{\micro\second}, allowing our \glspl{snn} to accurately classify more than \SI{85}{\kilo\nothing} images per second with a latency of \SI{8}{\micro\second}.

Moreover, we measured the system's power consumption.
When emulating the trained \gls{snn}, the full BrainScaleS-2 chip consumed approximately \SI{200}{\milli\watt}.
This figure included the current draw from the analog neuromorphic core, the plasticity processors, all surrounding periphery, and the high-speed communication links.
Combining this measurement with the above throughput results in an energy consumption of \SI{2.4}{\micro\joule} per classified image.

\paragraph{Efficiency through sparse spiking activity.}
A key advantage of \glspl{snn} is their sparse temporal spiking activity, which is presumed crucial for the power efficiency of the brain \citep{sterling2015principles}.
For similar reasons it is also important for larger neuromorphic systems and, particularly, in scenarios in which several chips cooperate by exchanging spikes over communication channels with limited bandwidth.

To ensure sparse activity on the BrainScaleS-2 system, we augmented the training loss by a regularization term
\begin{align}
	\mathcal{L_\mathrm{reg}} &= \rho_\text{b} \frac{1}{N_\text{H}} \sum_{i=1}^{N_\text{H}} \left( \sum_t S^\mathrm{H}_i[t] \right)^2 \label{eqn:burst_reg} \,,
\end{align}
with the strength parameter $\rho_\text{b}$, the hidden layer size $N_\text{H}$, and the corresponding hidden layer spike trains $S_i^\mathrm{H}$ \citep{zenke2021remarkable}.
We trained the above feed-forward \glspl{snn} for a range of different values $\rho_\text{b}$ and measured both their accuracy and average hidden layer spike counts.
All resulting network configurations were able to fit the training data with high accuracy (Fig.~\ref{fig:mnist_traces}E).
More importantly, they reached a constant test accuracy of \SI{97.2}{\percent} for activity levels down to approximately \num{20} hidden layer spikes per image.
When only using \num{10} spikes on average, we observed a slight decrease in performance.
At such low spike counts, the networks operated in a regime far from the rate coding limit and hence had to rely on individual spikes and their timing.

\begin{figure}[t!]
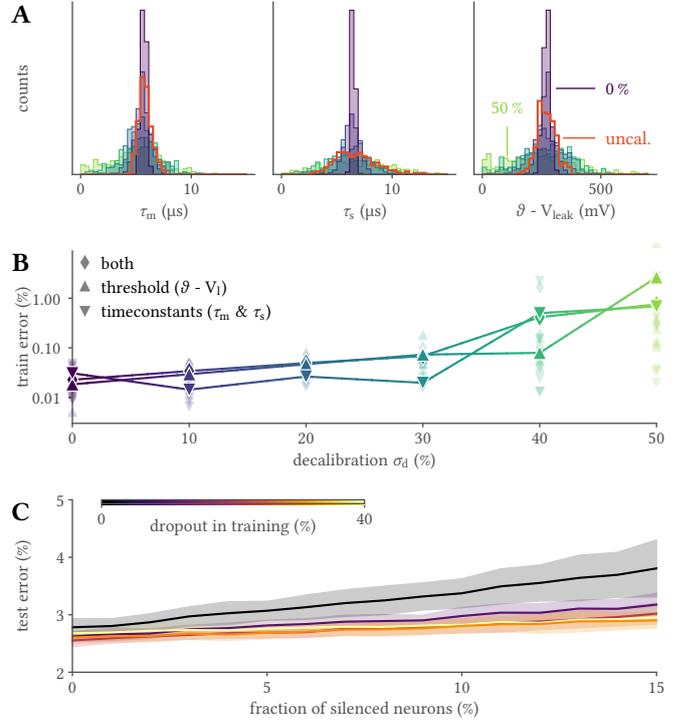

	\begin{center}
		\begin{tikzpicture}
			\node[anchor=north west,inner sep=0pt] (a) at (0.0,0) {\subimport{figures/decalibration/}{histograms.pgf}};
			\node[anchor=north west,inner sep=0pt] (b) at (0.0,-3.3) {\subimport{figures/decalibration/}{sweep.pgf}};
			\node[anchor=north west,inner sep=0pt] (c) at (0.0,-6.6) {\subimport{figures/blacklist/}{blacklist.pgf}};
			
			\node[inner sep=0pt] at ($(a.north west) + (0.11,-0.25)$) {\bfseries A};
			\node[inner sep=0pt] at ($(b.north west) + (0.11,-0.25)$) {\bfseries B};
			\node[inner sep=0pt] at ($(c.north west) + (0.11,-0.25)$) {\bfseries C};
		\end{tikzpicture}
	\end{center}
	\caption{
		\textbf{Self-calibration and robust performance on inhomogeneous substrates.}
	   	\textbf{(A)} Distribution of measured neuronal parameters for various degrees of decalibration in the range of \SIrange{0}{50}{\percent}.
	    		For this purpose, the analog circuits were deliberately detuned towards individual target values drawn from normal distributions of variable widths.
			Distributions for uncalibrated parameters are shown in red.
		\textbf{(B)} Despite assuming homogeneously behaving circuits in the computation graph, \gls{itl} training widely compensated the fixed-pattern deviations shown in panel A.
			For configurations with extreme mismatch, some networks suffered from dysfunctional states (e.g. leak-over-threshold).
		\textbf{(C)} When incorporating dropout regularization during training, networks become widely resilient to failure of hidden neurons.
	    }
	\label{fig:robustness}
\end{figure}

\paragraph{Self-calibration through \gls{itl} learning.}

The above results were obtained with a calibrated BrainScaleS-2 system in which the parameter deviations due to device-mismatch were largely compensated and the computation graph hence closely matched the emulated dynamics.
Nevertheless, a certain degree of residual mismatch remained.
To quantify whether and how well our \gls{itl} scheme self-calibrates the substrate during learning, we performed a series of additional experiments in which we deliberately decalibrated the system.
Specifically, we calibrated each neuron's time constants and threshold to individual target values.
These were drawn from normal distributions with a mean corresponding to the original calibration targets.
We generated parameter sets by varying their normalized standard deviations $\sigma_\text{d}$ in the range of \SIrange{0}{50}{\percent} (Fig.~\ref{fig:robustness}A).
This notably exceeded the mismatch present on an uncalibrated BrainScaleS-2 system.
To dissect the influence of poorly matching time constants and misaligned thresholds, we first detuned $\tau_\text{m,s}$ and $\vartheta - V_\text{leak}$ separately and finally all of these parameters at the same time.
Each of these experiments was repeated for five random seeds.

For each set of parameters, we trained the \gls{snn} on the neuromorphic system, still assuming ideal dynamics when constructing the computation graph as done previously.
In other words we explicitly ignored the introduced mismatch.
Nevertheless, learning performance was hardly affected by decalibration up to $\sigma_\text{d} = \SI{30}{\percent}$.
Beyond that point, error levels remained low but gradually increased.
At these levels of decalibration some configurations suffered from pathological network states, caused by some neurons having supra-threshold resting potentials.
Thus, even for mismatch levels far above the ones expected for BrainScaleS-2 and similar systems, \gls{itl} learning effectively self-calibrated the analog neuromorphic \glspl{snn}.

\paragraph{Training for robustness.}

We furthermore investigated the resilience of trained \glspl{snn} to defects occurring after deployment, e.g., failing neuron circuits.
To this end, we simulated neuronal death by artificially silencing randomly selected units in the hidden layer of the network after training.
As expected, performance deteriorated with an increasing fraction of disabled neurons (Fig.~\ref{fig:robustness}C).

However, when robustness was encouraged during training using dropout, the resilience to such neuronal failures was largely improved.
For networks trained with a dropout rate of \SI{40}{\percent} the test error increased by only \SI{10}{\percent} when silencing \SI{15}{\percent} of the hidden layer units.
In contrast, it grew by \SI{37}{\percent} when dropout was not used during training.

\begin{figure*}[t!]
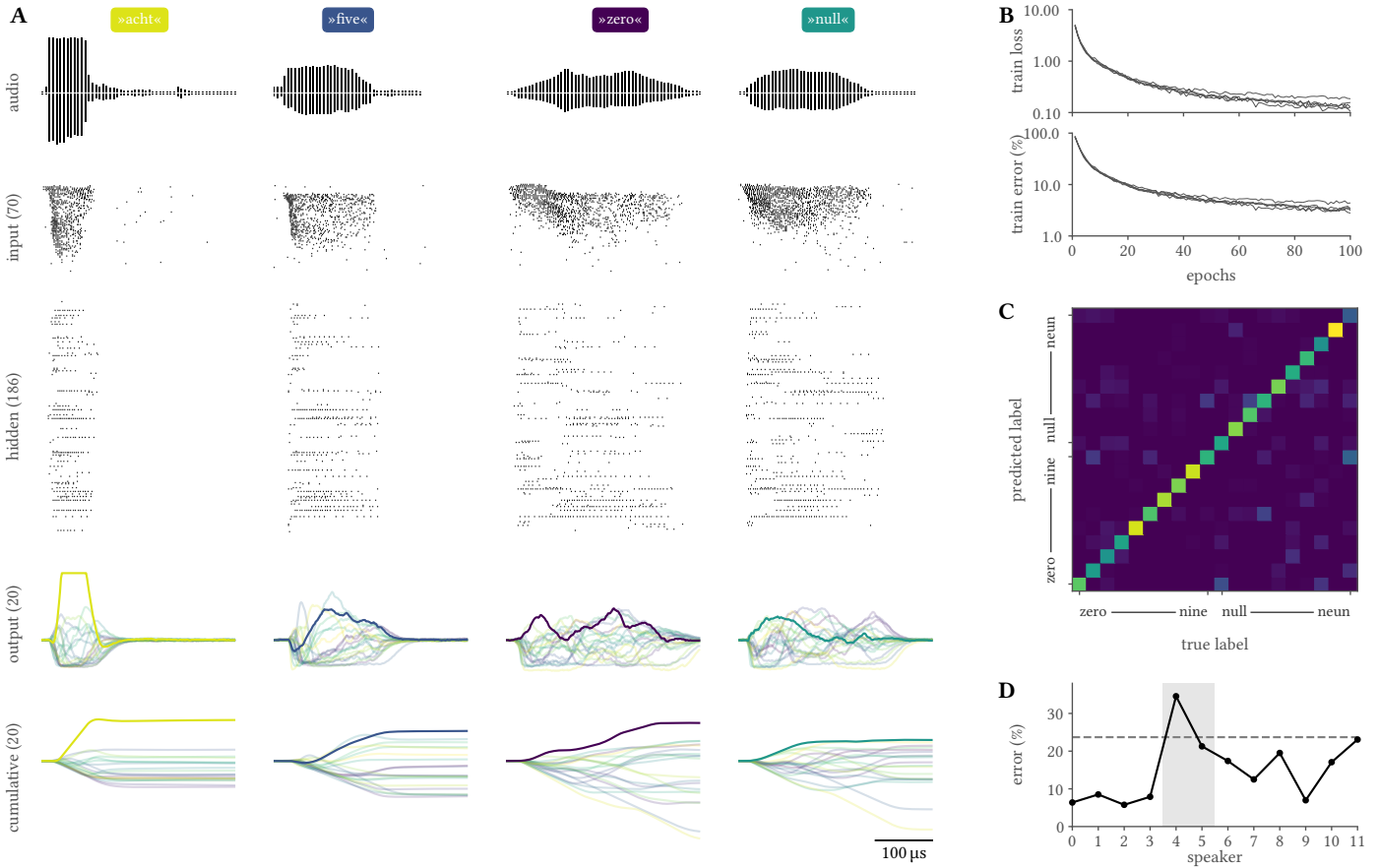

		\begin{tikzpicture}
			\node[anchor=north west,inner sep=0pt] (a) at (0.0,0) {\subimport*{figures/shd/}{spikes_traces.pgf}};
			\node[anchor=north east,inner sep=0pt] (b) at (\textwidth,0) {\subimport*{figures/shd-accuracy/}{accuracy.pgf}};
			\node[anchor=north east,inner sep=0pt] (c) at (\textwidth,-4.0) {\subimport*{figures/shd/}{confusion.pgf}};
			\node[anchor=north east,inner sep=0pt] (d) at (\textwidth,-9.1) {\subimport*{figures/shd/}{speaker.pgf}};
			
			\node[inner sep=0pt] at ($(a.north west) + ( 0.20,-0.15)$) {\small\bfseries A};
			\node[inner sep=0pt] at ($(b.north west) + ( 0.05,-0.15)$) {\small\bfseries B};
			\node[inner sep=0pt] at ($(c.north west) + ( 0.05,-0.15)$) {\small\bfseries C};
			\node[inner sep=0pt] at ($(d.north west) + ( 0.05,-0.15)$) {\small\bfseries D};
		\end{tikzpicture}
	\caption{
		\textbf{Classification of natural language with recurrent \glspl{snn} on BrainScaleS-2.}
		\textbf{(A)} Responses of a recurrent network when presented with samples from the \gls{shd} dataset.
			The input spike trains, originally derived from recordings of spoken digits (see illustrations), were reduced to 70 stimuli.
			The network was trained according to a sum-over-time loss based on the output units' membrane traces.
			For visualization purposes, we also show their cumulative sums.
		\textbf{(B)} Over 100 epochs of training, the network developed suitable representations as evidenced by a reduced training loss and error, here shown for five distinct initial conditions.
		\textbf{(C)} Classification performance varies across the twenty classes, especially since some of them exhibit phonemic similarities (»nine« vs. »neun«).
		\textbf{(D)} The trained network generalizes well on unseen data from most speakers included in the dataset.
			The discrepancy between training and overall test error (dashed line) arises from the composition of the dataset:
			\SI{81}{\percent} of the test set's samples stem from two exclusive speakers (highlighted in gray).
	    }
	\label{fig:shd}
\end{figure*}

\paragraph{Speech recognition with recurrent \acrshortpl{snn}.}

So far, our analysis was limited to tasks with short time horizons which can be readily solved using feed-forward networks.
But other tasks such as speech recognition or keyword spotting may require working memory and thus recurrent architectures.
On BrainScaleS-2, recurrent connectivity is readily supported by a flexible event router.
Further, recurrence is easily integrated into our \gls{itl} learning scheme by adding recurrent connections to Eq.~\ref{eqn:currents}.

With these modifications, we trained a network with 186~recurrently connected hidden neurons to classify the \gls{shd} dataset~\citep{cramer2019heidelberg}, which consists of spoken digits from »zero« to »nine« in both English and German, resulting in 20 classes total.
This dataset is a natural benchmark for \glspl{snn} due to its inherent temporal dimension.
Furthermore, it directly provides input spike trains and hence alleviates the need for additional preprocessing, which can confound comparison.
To feed the data into our system, we reduced their dimensionality by subsampling 70 out of the original 700 channels~(cf.~\nameref{sec:methods}).
The network was then trained by optimizing a sum-over-time loss $ \mathcal{L} = \operatorname{NLL} ( \operatorname{softmax} ( \operatorname{sum}_t \, V_i^\text{O}[t] ), y^\star ) $ (Fig.~\ref{fig:shd}A).
To prevent pathologically high firing rates, we employed homeostatic regularization during training.
Specifically, we added a regularizer of the form $\rho_\text{r} \cdot \max(0, \sum_{i,t} S_i[t] - \vartheta_\text{r})^2$, where $i$ and $t$ iterate over the hidden layer units and time steps, respectively, $\rho_\text{r}$ defines the regularization strength, and $\vartheta_\text{r}$ an activity threshold.

After 100 training epochs, the \gls{snn} reached \SI{96.6\pm0.5}{\percent} on the training data and \SI{76.2\pm1.3}{\percent} on the test set (Fig.~\ref{fig:shd}B, Table~\ref{tab:results}).
The large gap is presumably due to the nature of the dataset, which was designed to especially challenge a network's ability to generalize \citep{cramer2019heidelberg}.
The two languages included in the dataset exhibit classes with significant phonemic similarity (»nine« vs. »neun«), which are indeed harder to separate by the trained network (Fig.~\ref{fig:shd}C).
Most importantly, however, the test set consists to \SI{81}{\percent} of two speakers that are not part of the training set and result in higher classification error rates (Fig.~\ref{fig:shd}D).
To improve generalization performance, we employed data augmentation.
For this purpose, we stochastically shifted events to neighboring input channels drawn from a normal distribution centered around their original channel (cf. \nameref{sec:methods}).
This indeed improved the test performance to \SI{80.6\pm1.0}{\percent}.

As for the feed-forward network, we additionally trained and evaluated the \gls{snn} in an equivalent software-only implementation.
Without augmentation it reached an accuracy of \SI{71.2 \pm 0.3}{\percent} --- far below the corresponding hardware results.
At the same time, the software simulation was able to perfectly fit the training data, which was not achieved on \mbox{BrainScaleS-2}.
We hypothesize that this discrepancy resulted from the intrinsic stochasticity of the analog substrate, which was propagated and amplified by the network's recurrent dynamics and acted as a form of regularization.
To test this idea, we applied data augmentation to the simulation.
Indeed, this change resulted in a improved test accuracy of \SI{79.9 \pm 0.7}{\percent}, close to the performance of BrainScaleS-2 under equivalent conditions.
Similarly, augmentation closed the gap between software and hardware training accuracy, providing further support for our hypothesis.
Thus, our work suggests that intrinsic analog device noise could act as an efficient regularizer.
Importantly, our findings illustrate that the flexibility of \gls{itl} learning also applies to the realm of recurrent \gls{snn} trained on challenging speech processing problems.

\section*{Discussion}

\begin{table*}[t!]
	\centering
	\fontsize{8.5}{10}\selectfont
	\caption{Comparison of MNIST benchmark results across neuromorphic platforms.}
	\label{tab:comparison}

	\begin{tabular}{clllrrrrr}
	\toprule
		& platform        & reference					& architecture
		& node                      & accuracy			& energy/inference
		& inferences/s			& latency			\\
	\midrule                                                                                                                            
	\multirow{8}{*}{\rotatebox[origin=c]{90}{digital}} %
		& SpiNNaker       & Stromatias et al., 2015 \citep{stromatias2015robustness}	& 784-500-500-10		& \SI{130}{\nano\meter}     & \SI{95.0}{\percent}	& -\hphantom{\,ms}\rlap{\textsuperscript{iii}}				& -\hphantom{\,k}\rlap{\textsuperscript{iii}}	& -\hphantom{ms} 				\\
		& TrueNorth       & Esser et al., 2016 \citep{esser2016convolutional}	& CNN				& \SI{28}{\nano\meter}      & \SI{99.4}{\percent}	& \SI{108.0}{\micro\joule}				& \SI{1}{\kilo\nothing}		& -\hphantom{ms}				\\
		& Intel           & Chen et al., 2019 \citep{chen20184096}			& 236-20	& \SI{10}{\nano\meter}      & \SI{88.0}{\percent}	& \SI{1.0}{\micro\joule}				& \SI{6.25}{\kilo\nothing}	& -\hphantom{ms}				\\
		& Intel           & Chen et al., 2019 \citep{chen20184096}			& 784-1024-512-10		& \SI{10}{\nano\meter}      & \SI{98.2}{\percent}	& \SI{12.4}{\micro\joule}				& -\hphantom{\,k}				& -\hphantom{ms}				\\
		& Intel           & Chen et al., 2019 \citep{chen20184096}			& 784-1024-512-10		& \SI{10}{\nano\meter}      & \SI{97.9}{\percent}	& \SI{1.7}{\micro\joule}				& -\hphantom{\,k}				& -\hphantom{ms}				\\
		& MorphIC         & Frenkel et al., 2019 \citep{frenkel2019morphic}		& 784-500-10\rlap{\textsuperscript{i}}			& \SI{65}{\nano\meter}      & \SI{97.8}{\percent}	& \SI{205}{\micro\joule}				& -\hphantom{\,k}				& -\hphantom{ms}				\\
		& MorphIC         & Frenkel et al., 2019 \citep{frenkel2019morphic}		& 784-500-10\rlap{\textsuperscript{i}}			& \SI{65}{\nano\meter}      & \SI{95.9}{\percent}	& \SI{21.8}{\micro\joule}				& \SI{250}{\nothing}\hphantom{k}& -\hphantom{ms}				\\ %
		& SPOON           & Frenkel et al., 2020 \citep{frenkel202028}			& CNN				& \SI{28}{\nano\meter}      & \SI{97.5}{\percent}	& \SI{0.3}{\micro\joule}\rlap{\textsuperscript{ii}}	& -\hphantom{\,k}				& \SI{117}{\micro\second}	\\
                                                                                                   
	\midrule                                                                                                                            
		\multirow{3}{*}{\rotatebox[origin=c]{90}{analog}} %
		& BSS-1           & Schmitt et al., 2017 \citep{schmitt2017neuromorphic}		& 100-15-15-5			& \SI{180}{\nano\meter}     & \SI{95.0}{\percent}	& -\hphantom{ms}\rlap{\textsuperscript{iii}}							& \SI{10}{\kilo\nothing}	& -\hphantom{ms}				\\
		& BSS-2           & Göltz et al., 2021 \citep{goltz2019fast}			& 256-246-10			& \SI{65}{\nano\meter}      & \SI{96.9}{\percent}	& \SI{8.4}{\micro\joule}				& \SI{21}{\kilo\nothing}	& \SI{< 10}{\micro\second}	\\
		& BSS-2           & this work					& 256-246-10			& \SI{65}{\nano\meter}      & \SI{97.6}{\percent}	& \SI{2.4}{\micro\joule}				& \SI{85}{\kilo\nothing}	& \SI{8}{\micro\second}		\\
	\bottomrule
	\end{tabular}

	\addtabletext{\sffamily\textsuperscript{i} Segmented input and hidden layers. \textsuperscript{ii} Based on pre-silicon estimates. \textsuperscript{iii} Estimates were given by Pfeiffer et al., 2018~\citep{pfeiffer2018deep}.}
\end{table*}

We have developed a general \gls{itl} learning method for recurrent and multi-layer \glspl{snn} on analog neuromorphic substrates and demonstrated its capabilities on BrainScaleS-2.
The combination of surrogate gradients with \gls{itl} training -- facilitated by the massively parallel digitization of analog membrane potentials -- allowed us to tie on recent achievements in the field of \gls{snn} optimization and bring them to an analog substrate.
This allowed us to achieve state-of-the-art classification accuracies on multiple benchmark problems, comparable to equivalent software simulations.
During training, our framework automatically corrected for device mismatch and thus abolished the need for explicit calibration.
The resulting \glspl{snn} exhibited spatially and temporally sparse activity patterns and could, furthermore, be optimized for resilience to neuron failure.
Ultimately, our method allowed us to exploit BrainScaleS-2 for low-latency neuromorphic inference at high throughput and a low energy footprint.

Most current neuromorphic systems are fully digital and typically allow to simulate software trained models without performance loss \citep{esser2016convolutional, frenkel202028}.
This approach is flexible with regard to the \gls{snn} training schemes used \citep{bohte_error-backpropagation_2002, ruckauer2019closing, zambrano2017efficient, pfeiffer2018deep, buchel2021supervised, hunsberger2015spiking, lee2016training, neftci2019surrogate, mostafa_supervised_2018, bellec2020solution, NEURIPS2018_185e65bc}.
Still, research on analog and mixed-signal substrates is essential to turn recent advances in material science into efficient neuromorphic processors \citep{roy2019towards, markovic2020physics, joshi2020accurate, dalgaty2021situ}.
Key emerging technologies like memristors are ideal candidates for long-term memory storage in neuromorphic systems.
However, these respective components are intrinsically analog and subject to drift and manufacturing variability.
These imperfections lead to reduced performance when loading software trained models onto the analog substrate. While several studies approached this problem by optimizing for additional robustness during training \citep{buchel2021supervised,wright2021deep}, 
these techniques are intrinsically limited.
Since mature on-chip training solutions are not yet available, \gls{itl} learning has emerged as a good compromise that efficiently takes device-specific non-idealities and heterogeneity into account \citep{schmitt2017neuromorphic, goltz2019fast}.
However, previous work relied on rate-based or time-to-first-spike coding schemes.
Here, we expanded \gls{itl} techniques into the realm of surrogate gradient learning which flexibly interpolates between rate- and timing-based coding schemes on multi-layer and recurrent architectures, thereby simultaneously improving performance and energy efficiency, while also being conducive for fast inference \citep{davidson2021comparison}.

Comparing the performance of neuromorphic \gls{snn} implementations is an intricate task in itself, starting with a lack of standardized benchmarks \citep{davies2019benchmarks, cramer2019heidelberg}.
When aspiring to sound the spectrum of different neuromorphic architectures and \gls{snn} coding schemes, one furthermore stumbles upon inhomogeneous standards of determining a system's energy consumption, ranging from pre-silicon estimates of a neuromorphic core's current draw to full-system lab measurements.
We, nevertheless, attempted to contrast our findings with results from previous studies on both digital as well as analog systems (Table~\ref{tab:comparison}).
Although lacking the essence of temporal spike-based information processing, we considered the MNIST dataset due to its widespread adoption.

Our model on BrainScaleS-2 performed competitively in all metrics and -- in terms of accuracy -- was surpassed only by much larger or convolutional networks.
When considering the energy footprint, BrainScaleS-2 reached values only outperformed by optimized architectures fabricated in much smaller and hence more efficient technology nodes \citep{frenkel202028,chen20184096}.
In comparsion to other neuromorphic systems, we were able to set new benchmarks in terms of throughput and classification latency.

In summary, our work shows how learning can efficiently compensate for device-specific imperfections, thereby allowing us to employ analog neuromorphic substrates for complex, energy-efficient, and ultra-low latency information processing.
Importantly, it also is the first step toward future on-chip learning algorithms that could even take advantage of such device heterogeneity \citep{perez-nieves2021neural}. 
Thus our work gives a glimpse of how powerful learning algorithms will empower future neuromorphic technologies.

\section*{Materials and methods}
\label{sec:methods}

\paragraph{Software environment}
Our training framework was based on PyTorch's auto-differentiation library~\citep{NEURIPS2019_9015}.
It furthermore builds upon the BrainScaleS-2 software stack to configure the neuromorphic system and execute the experiments \citep{muller2020extending}.

\paragraph{Input coding}
For MNIST we scaled down the dataset to 16$\times$16 pixels by first discarding the two outermost rows and scaling the remaining pixels.
The images were then converted to spikes by interpreting the normalized pixel grayscale values $x_i$ as input currents to \gls{lif} neuons.
Strong enough currents trigger a spike at time $t_i = \tau_\mathrm{in} \log x_i/(x_i-\vartheta_\mathrm{in})$, where $\tau_\mathrm{in}$ denotes the input units time constant and $\vartheta_\mathrm{in}$ its threshold.

Since the \gls{shd} dataset is provided in form of input spike times, a custom conversion was not required.
For \gls{shd} we reduced the original 700 input channels by subsampling.
Specifically, we omitted the first 70 and then retained every ninth input unit.
The time dimension was scaled by a factor of 2000 to account for the system's acceleration factor of of 1000 and further shorten the experiment duration to reduce the computation burden on the host system.
When employing data augmentation, a spike originally originating from input channel $i$ was reassigned to a neighboring channel drawn from $\mathcal{N}(\mu=i, \sigma)$.
This augmentation was applied prior to downsampling the inputs.

\paragraph{Initialization}
We used Kaming's initialization \citep{he2015delving} for both the hidden and output layer weights.
Specifically, weights were drawn from a normal distribution with zero mean and a standard deviation of $\hat\sigma_w / \sqrt{N_\text{H,L}}$.

\paragraph{Weight scaling}
Weight values had to be scaled, rounded, and cropped to the neuromorphic system's weight resolution of \SI{7}{\bit} signed integers resulting from merging two \SI{6}{\bit} synapse circuits.
The exact scaling took into account analog bias currents and other technical parameters.
Due to the absence of a threshold for the non-spiking output layer, its membrane traces could be scaled arbitrarily.

For the MNIST classification, we adopted a dynamic weight scaling for the output weights by aligning the largest absolute weight value as represented in software to the maximum weight possible on the substrate.

\noindent
\begin{table}[h]
	\newcommand\common[1]{\multicolumn{3}{c}{#1}}
	\caption{Parameters for the neuromorphic substrate and learning framework.}
	\label{tab:params}
	\begin{center}
	\fontsize{8.5}{10}\selectfont
		\begin{tabular}{l>{\raggedleft}p{1.55cm}@{\hskip 0.3em}c@{\hskip 0.3em}p{1.55cm}}
		\toprule
		parameter & \multicolumn{3}{c}{value (MNIST / SHD)} \\
		\midrule
		difference threshold-leak $\vartheta - V_\text{leak}$ & \common{\SI{270\pm15}{\milli\volt}} \\
		membrane time constant $\tau_\text{m}$ & \SI{5.7\pm0.3}{\micro\second} & / & \SI{10.0\pm0.3}{\micro\second} \\
			$\quad$ in computation graph & \SI{6.0}{\micro\second} & / & \SI{10.0}{\micro\second} \\
		synaptic time constant $\tau_\text{s}$ & \SI{6}{\micro\second} & / & \SI{10}{\micro\second} \\
		\midrule
		input unit time constant $\tau_\text{in}$ & \SI{8}{\micro\second} & / & -- \\
		input unit threshold $\vartheta_\text{in}$ & \num{0.2} & / & -- \\
		\midrule
		surrogate gradient steepness $\beta$ & \common{\num{50}} \\
		learning rate $\eta$ & \common{\num{1.5e-3}} \\
		learning rate decay per epoch $\gamma_\eta$ & \num{0.03} & / & \num{0.025} \\
		amplitude regularization strength $\rho_\text{a}$ & \num{4e-4} & /&  -- \\
		burst regularization strength $\rho_\text{b}$ & \num{0.005} & /&  -- \\
		rate regularization strength $\rho_\text{r}$ & -- & /&  \num{0.6e-3} \\
		rate regularization threshold $\vartheta_\text{r}$ & -- & / & \num{600} \\
		\midrule
		time step/sample period $\Delta t$ & \common{\SI{1.7}{\micro\second}} \\
		weight initialization spread $\hat\sigma_w $ & \common{\SI{0.24}{\nothing}} \\
		\bottomrule
	\end{tabular}
	\end{center}
\end{table}

\section*{Contributions}
B. Cramer, S. Billaudelle, and F. Zenke conceived the work.
B. Cramer and S. Billaudelle implemented the software, performed the experiments, and analyzed the data.
S. Kanya and A. Leibfried contributed to the implementation.
J. Schemmel is the lead designer and architect of the BrainScaleS-2 neuromorphic system.
A. Grübl, V. Karasenko, C. Pehle, K. Schreiber, and Y. Stradmann contributed to the design of the BrainScaleS-2 system.
J. Weis contributed the calibration routines.
B. Cramer, S. Billaudelle, and F. Zenke wrote the manuscript with input from the other authors.

\section*{Acknowledgements}
We express our gratitude towards O.\ Breitwieser, C.\ Mauch, E.\ Müller, and P.\
Spilger for their work on the software environment, 
B.\ Kindler, F.\ Kleveta, and S.\ Schmitt for their helpful support, 
A.\ Baumbach for his valuable feedback during the early commissioning phase of the system, 
and J.\ Göltz and L.\ Kriener for helpful discussions. 
We thank the whole Electronic Vision(s) group for the inspirational work environment.

This research has received funding from the European Union’s Horizon 2020 research and innovation programme under grant agreement Nos. 720270, 785907 and 945539 (Human Brain Project, HBP).
This work was supported by the Novartis Research Foundation.

\bibliography{superspike}

\end{document}